\pdfoutput=1

\documentclass[11pt]{article}
\usepackage{authblk}
\usepackage[]{acl}

\usepackage{times}
\usepackage{latexsym}

\usepackage[T1]{fontenc}

\usepackage[utf8]{inputenc}

\usepackage{microtype}
\usepackage{graphicx}
\usepackage{subcaption}
\usepackage{multirow}

\makeatletter
\def\thickhline{%
  \noalign{\ifnum0=`}\fi\hrule \@height \thickarrayrulewidth \futurelet
   \reserved@a\@xthickhline}
\def\@xthickhline{\ifx\reserved@a\thickhline
               \vskip\doublerulesep
               \vskip-\thickarrayrulewidth
             \fi
      \ifnum0=`{\fi}}
\makeatother

\newlength{\thickarrayrulewidth}
\title{ChatGPT as Data Augmentation for Compositional Generalization: A Case Study in Open Intent Detection}

\author[2,3]{Yihao Fang}
\author[1,2]{Xianzhi Li}
\author[3]{Stephen W. Thomas}
\author[1,2]{Xiaodan Zhu}
\affil[1]{Department of Electrical and Computer Engineering, Queen's University}
\affil[2]{Ingenuity Labs Research Institute, Queen's University}
\affil[3]{Smith School of Business, Queen’s University \authorcr
\{yihao.fang, 21xl17, stephen.thomas, xiaodan.zhu\}@queensu.ca}

\usepackage{fancyhdr}

\fancypagestyle{Proceedings}{\fancyhf{} \fancyfoot[C]{Proceedings of the Joint Workshop of the 5th Financial Technology and Natural Language Processing (FinNLP) and
2nd Multimodal AI For Financial Forecasting (Muffin), Macao, August 20, 2023.} }

\begin{document}
\thispagestyle{Proceedings}
\maketitle
\begin{abstract}
Open intent detection, a crucial aspect of natural language understanding, involves the identification of previously unseen intents in user-generated text. Despite the progress made in this field, challenges persist in handling new combinations of language components, which is essential for compositional generalization. In this paper, we present a case study exploring the use of ChatGPT as a data augmentation technique to enhance compositional generalization in open intent detection tasks. We begin by discussing the limitations of existing benchmarks in evaluating this problem, highlighting the need for constructing datasets for addressing compositional generalization in open intent detection tasks. By incorporating synthetic data generated by ChatGPT into the training process, we demonstrate that our approach can effectively improve model performance. Rigorous evaluation of multiple benchmarks reveals that our method outperforms existing techniques and significantly enhances open intent detection capabilities. Our findings underscore the potential of large language models like ChatGPT for data augmentation in natural language understanding tasks.
\end{abstract}

\section{Introduction}

Open intent detection, a key component of natural language understanding, aims to identify previously unseen intents in user-generated text. This task is of paramount importance for a wide range of applications, such as conversational AI systems, where the ability to recognize new intents can substantially improve the user experience. Although the field has made significant strides in recent years, a major challenge remains in addressing compositional generalization, which refers to the capability of models to handle unseen combinations of language components. This capability is essential for the successful deployment of AI systems in real-world scenarios, where users may express intent in unforeseen ways.

In this paper, we present a case study that investigates the potential of ChatGPT, a state-of-the-art large language model, as a data augmentation technique for enhancing compositional generalization in open intent detection tasks. Our study begins by identifying the shortcomings of existing benchmarks in evaluating this problem, which underscores the need for the development of datasets tailored to assess compositional generalization in open intent detection tasks.

To address this issue, we leverage ChatGPT to generate synthetic data that is then incorporated into the training process. By doing so, we aim to improve the model's ability to recognize new combinations of language components, thereby enhancing its open intent detection capabilities. Through rigorous evaluation of multiple benchmarks, we demonstrate that our proposed method outperforms existing techniques and leads to significant performance improvements.

Our findings highlight the potential of large language models, such as ChatGPT, for data augmentation in natural language understanding tasks. This case study offers valuable insights into the development of more effective dialogue systems capable of handling a wider range of user intents and fostering better human-computer interactions.

\begin{figure*}[ht]
\begin{subfigure}{.5\textwidth}
  \centering
  \includegraphics[width=.9\linewidth]{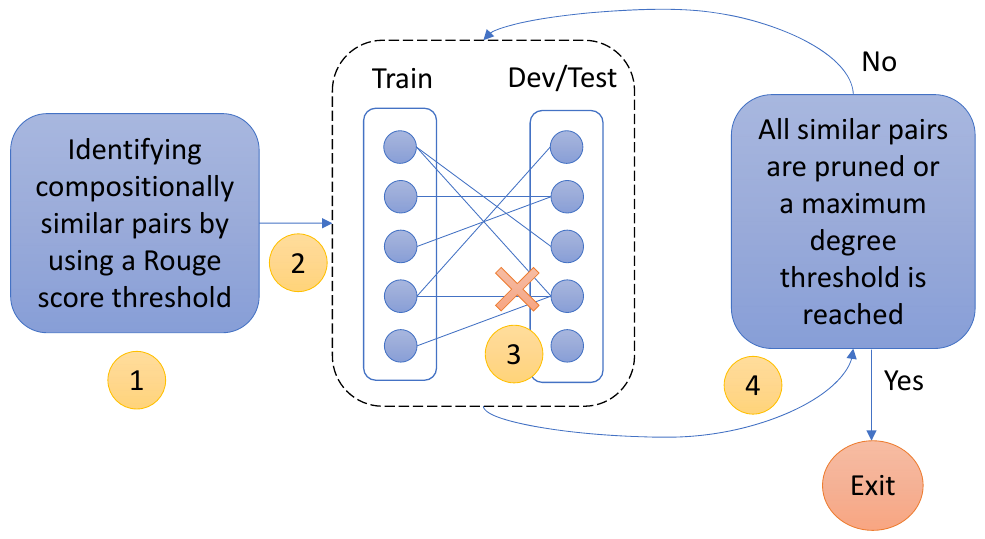}  
  \caption{Dataset construction}
  \label{fig:dataset_construction}
\end{subfigure}
\begin{subfigure}{.5\textwidth}
  \centering
  \includegraphics[width=.9\linewidth]{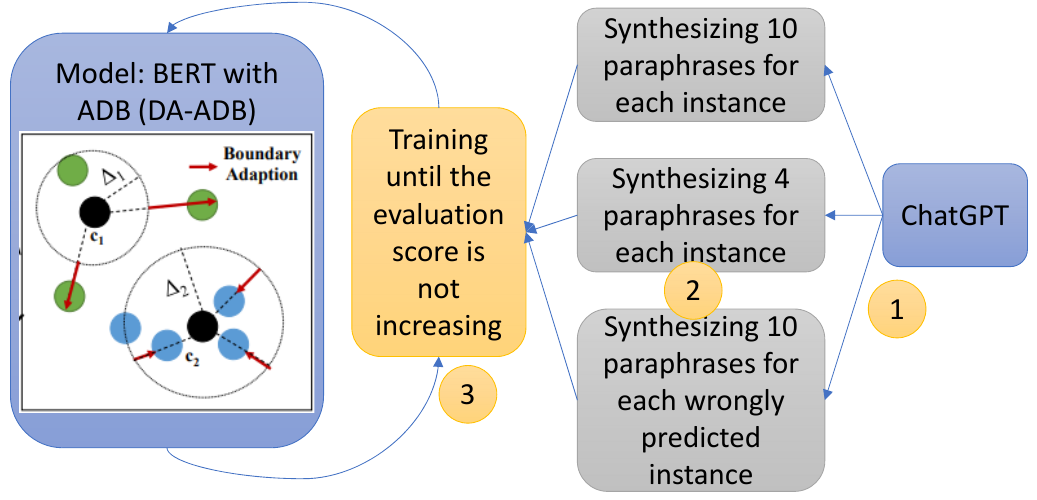}  
  \caption{ChatGPT data augmentation}
  \label{fig:chatgpt_data_augmentation}
\end{subfigure}
\caption{a-1) Compositionally-similar utterance instances are identified by a Rouge score threshold. a-2) An undirect graph is created by connecting compositionally-similar instances with edges. a-3) Node (instance) degrees are counted in the undirected graph and highest-degree nodes and their edges are first pruned. a-4) This process iterates until all similar pairs are pruned or a maximum degree threshold is reached. b-1) Paraphrases are generated by ChatGPT for utterances in the training dataset. b-2) Paraphrases are incorporated into the BERT training process through three different strategies. b-3) The training process iterates until the evaluation score (accuracy) is not increasing. }
\end{figure*}

Our primary contributions to the literature include:
\begin{itemize}
\item
Dataset Construction for Compositional Generalization: We construct compositionally diverse subsets derived from existing open intent detection benchmark datasets. 
\item
ChatGPT Data Augmentation: We propose using ChatGPT to generate paraphrases of training dataset instances, thereby enhancing model generalization and performance on unseen compositions. 
\item
We evaluate three different strategies for incorporating ChatGPT-generated paraphrases into the training process of BERT \cite{devlin-etal-2019-bert} with ADB \cite{Zhang:21} (DA-ADB \citealp{DA-ADB}).
\end{itemize}


The rest of the paper is organized as follows: Section~\ref{sec:related_work} provides a background on open intent classification and reviews related work. Section~\ref{sec:methodology} describes our proposed method in detail. Section~\ref{sec:experiments} presents the experimental setup, results, and analysis. Finally, Section~\ref{sec:conclusion} concludes the paper and suggests directions for future research.

\section{Related Work}
\label{sec:related_work}
Open intent classification is an important problem in natural language understanding and dialogue systems, aiming to identify known intents and detect unseen open intents using only the prior knowledge of known intents. Several recent studies have explored various techniques for addressing this challenging task.

One line of research involves aligning representation learning with scoring functions. For instance, the unified neighbourhood learning framework (UniNL) was proposed to detect OOD intents by designing a KNCL objective for representation learning and introducing a KNN-based scoring function for OOD detection \cite{mou-etal-2022-uninl}. Another study proposed a unified K-nearest neighbour contrastive learning framework for OOD intent discovery, which focuses on inter-class discriminative features and alleviates the in-domain overfitting problem \cite{mou-etal-2022-watch}.

Another direction focuses on learning discriminative representations and decision boundaries for open intent detection. 
The Deep Open Intent Classification with Adaptive Decision Boundary (\textbf{ADB}) method learns an adaptive spherical decision boundary for each known class, balancing both the empirical risk and the open space risk without requiring open intent samples or modifying the model architecture \cite{Zhang:21}. Similarly, the \textbf{DA-ADB} framework successively learns distance-aware intent representations and adaptive decision boundaries for open intent detection by leveraging distance information and designing a loss function to balance empirical and open space risks \cite{DA-ADB}. 

In summary, various methods have been proposed to address the challenges associated with detecting unseen intents. However, none of them have explored compositional generalization in open intent detection tasks. We highlight the need for constructing datasets and leverage ChatGPT to generate synthetic data to address this problem. Our proposed method in detail is given in the following section.

\section{Methodology}
\label{sec:methodology}
\subsection{Dataset Construction for Compositional Generalization}
The construction of the dataset starts with identifying compositionally-similar utterance instances by utilizing a Rouge score threshold (Figure~\ref{fig:dataset_construction}). The Rouge score is a widely-used metric for evaluating the similarity between a pair of text sequences by comparing the number of overlapping n-grams \cite{lin-2004-rouge}. By setting a threshold value, instances with Rouge scores above this threshold are deemed to be compositionally similar, allowing for the effective detection of instances with a high degree of overlap in content or structure. 

Once the compositionally-similar utterance instances are identified, an undirected graph is created by connecting these instances with edges. In this graph, each node represents an instance, and an edge is drawn between two nodes if their corresponding instances are compositionally similar according to the Rouge score threshold. This representation allows for a better understanding of the relationships between the instances, making it easier to discern patterns and outliers in the data. Furthermore, the graph-based approach facilitates the efficient pruning of highly similar instances in subsequent steps.

To refine the dataset and ensure maximum diversity, the highest-degree nodes and their connecting edges are first pruned. In this context, the degree of a node refers to the number of edges connected to it. By pruning the highest-degree nodes, the instances with the most similarities to other instances are removed from the dataset. This process iterates until all similar pairs have been pruned or a maximum degree threshold is reached. The result is a dataset with a high degree of diversity and helps to access the compositional generalizability of the model trained on this dataset.

The aforementioned approach is utilized on three open intent detection benchmark datasets: \textbf{Banking} \cite{casanueva-etal-2020-efficient}, \textbf{OOS} \cite{larson-etal-2019-evaluation} and \textbf{StackOverflow} \cite{xu-etal-2015-short}, resulting in three compositionally diverse subsets derived from these datasets, namely \textbf{Banking\_CG},  \textbf{OOS\_CG}, and \textbf{StackOverflow\_CG}. (Refer to Appendix~\ref{sec:appendix_dataset_construction_in_detail} for dataset construction in detail.) 

\subsection{ChatGPT Data Augmentation}
The training process involves generating paraphrases for utterances in the training dataset using ChatGPT (Figure~\ref{fig:chatgpt_data_augmentation}). This paraphrasing approach aids in enhancing the model's understanding of language by providing alternative compositions of the same meaning. The incorporation of these paraphrases into the training process not only improves the generalizability of the model but also leads to better performance on unseen compositions. (Refer to Appendix~\ref{sec:appendix_examples_of_chatgpts_paraphrases} for ChatGPT's paraphrases in detail.)

To effectively integrate paraphrases into the training process of BERT \cite{devlin-etal-2019-bert} with \textbf{ADB} (\textbf{DA-ADB}), three different strategies are evaluated. The first strategy involves synthesizing $10$ paraphrases for each instance in the dataset (\textbf{GPTAUG-F10}), while the second strategy generates $4$ paraphrases for each instance (\textbf{GPTAUG-F4}). The third strategy, on the other hand, focuses on instances that the model predicts incorrectly at the current iteration and synthesizes $10$ paraphrases for each of these instances (\textbf{GPTAUG-WP10}). This targeted approach aims to help address specific weaknesses in the model's understanding. The training process iterates through these strategies until the evaluation score, such as accuracy, no longer exhibits any improvement. This iterative process ensures that the model continues to refine its understanding of language by learning from the generated paraphrases, ultimately resulting in a more robust and capable BERT model.

\section{Experiments}
\label{sec:experiments}
\subsection{Experimental Setup}

In our experimental setup, we have extended the TEXTOIR platform \cite{zhang-etal-2021-textoir}, a toolkit that integrates a variety of state-of-the-art algorithms for open intent detection, to conduct our experiments. To ensure fair comparisons across all tests, we employed the pre-trained BERT-base model from Hugging Face \cite{wolf-etal-2020-transformers} as the foundation of our approach. The optimization of the BERT model with ADB (DA-ADB) was carried out using Python and the PyTorch framework \cite{paszke2019pytorch} and executed on NVIDIA RTX 2080 TI GPUs for computational efficiency.

\subsection{Results and Analysis}

\begin{table*}
\centering
\small
\caption{Performance of our ChatGPT augmentation approaches (GPTAUG-F4, GPTAUG-F10, and GPTAUG-WP10) and the baselines (ADB and DA-ADB). The best results among each setting are bolded. All results are an average of $10$ runs using $10$ different seed numbers considering that the selection of known intents is a pseudo-random process. (Refer to Appendix~\ref{sec:appendix_experimental_results_in_detail} in more detail.)}
\label{tab:experimental_results}
\begin{tabular}{|l|l|p{0.035\linewidth}p{0.035\linewidth}p{0.035\linewidth}p{0.035\linewidth}|p{0.035\linewidth}p{0.035\linewidth}p{0.035\linewidth}p{0.035\linewidth}|p{0.035\linewidth}p{0.035\linewidth}p{0.035\linewidth}p{0.035\linewidth}|}
\thickhline
&&\multicolumn{4}{|c|}{\bf Banking\_CG}&\multicolumn{4}{|c|}{\bf OOS\_CG}&\multicolumn{4}{|c|}{\bf StackOverflow\_CG}\\
\cline{3-14}
&\bf Methods&\bf F1-IND&\bf F1-OOD&\bf F1-All&\bf Acc-All&\bf F1-IND&\bf F1-OOD&\bf F1-All&\bf Acc-All&\bf F1-IND&\bf F1-OOD&\bf F1-All&\bf Acc-All\\
\hline
\multirow{8}{*}{25\%}&ADB&53.49&81.10&54.87&72.31&49.13&90.65&50.19&83.96&58.37&79.04&61.82&71.35\\
&DA-ADB&53.33&\bf 86.15&54.97&\bf 78.43&38.27&91.70&39.64&85.45&\bf 62.32&\bf 84.84&\bf 66.08&\bf 77.68\\
\cline{2-14}
&ADB+GPTAUG-F4&56.73&83.37&58.06&75.51&53.54&91.93&54.52&86.23&60.94&82.62&64.55&75.32\\
&ADB+GPTAUG-F10&\bf57.58&84.04&\bf 58.90&76.46&\bf 54.26&\bf 92.07&\bf 55.23&\bf 86.48&58.99&80.34&62.55&72.66\\
&ADB+GPTAUG-WP10&50.04&70.47&51.06&61.47&48.03&88.57&49.07&80.83&51.62&63.54&53.60&56.17\\
&DA-ADB+GPTAUG-F4&54.58&84.82&56.09&77.11&43.98&91.85&45.21&85.89&59.97&78.03&62.98&70.79\\
&DA-ADB+GPTAUG-F10&53.52&84.00&55.04&76.00&44.20&91.74&45.42&85.70&59.82&75.31&62.40&70.10\\
&DA-ADB+GPTAUG-WP10&54.72&82.89&56.13&74.38&43.18&91.55&44.42&85.33&54.61&64.87&56.32&59.22\\
\hline\hline
\multirow{8}{*}{50\%}&ADB&59.93&69.63&60.18&65.38&52.32&83.99&52.73&75.66&71.45&76.14&71.88&73.58\\
&DA-ADB&54.57&74.45&55.08&67.77&33.66&83.31&34.31&73.37&\bf 75.97&\bf 81.75&\bf 76.49&\bf 79.14\\
\cline{2-14}
&ADB+GPTAUG-F4&\bf 62.55&73.20&\bf 62.83&69.24&55.36&85.39&55.76&78.07&71.58&77.03&72.08&74.37\\
&ADB+GPTAUG-F10&62.28&73.23&62.56&69.36&\bf 55.40&\bf 85.57&\bf 55.80&\bf 78.44&70.97&77.56&71.57&74.52\\
&ADB+GPTAUG-WP10&59.87&61.11&59.90&60.27&53.25&83.06&53.64&74.61&67.04&64.13&66.78&65.37\\
&DA-ADB+GPTAUG-F4&57.06&\bf 74.67&57.52&\bf 69.41&38.85&84.00&39.45&74.96&72.28&74.07&72.44&73.88\\
&DA-ADB+GPTAUG-F10&56.52&74.42&56.98&69.09&39.02&83.94&39.61&74.90&70.32&74.78&70.72&73.27\\
&DA-ADB+GPTAUG-WP10&58.63&70.33&58.93&65.30&40.26&83.99&40.84&74.81&69.91&64.58&69.43&66.65\\
\hline\hline
\multirow{8}{*}{75\%}&ADB&64.30&53.36&64.12&62.82&53.87&76.24&54.07&68.33&76.13&61.56&75.22&71.58\\
&DA-ADB&54.74&52.46&54.70&56.94&29.58&71.76&29.96&59.91&\bf 78.57&\bf 65.80&\bf 77.77&\bf 74.51\\
\cline{2-14}
&ADB+GPTAUG-F4&\bf 66.65&\bf 54.82&\bf 66.45&\bf 64.89&\bf 55.99&\bf 77.36&\bf 56.18&\bf 70.70&75.72&61.68&74.84&71.08\\
&ADB+GPTAUG-F10&66.22&54.61&66.02&64.54&55.64&77.04&55.83&70.54&75.35&61.15&74.46&70.61\\
&ADB+GPTAUG-WP10&65.22&47.98&64.93&62.22&54.91&75.78&55.10&67.97&73.86&49.92&72.37&67.67\\
&DA-ADB+GPTAUG-F4&55.87&51.59&55.80&57.67&33.50&72.40&33.85&61.81&76.87&60.79&75.86&71.51\\
&DA-ADB+GPTAUG-F10&54.86&50.66&54.78&56.65&33.81&72.48&34.15&62.04&73.65&54.81&72.48&68.10\\
&DA-ADB+GPTAUG-WP10&62.31&53.33&62.15&61.87&40.21&74.07&40.51&64.11&77.77&60.62&76.70&72.52\\
\thickhline
\end{tabular}
\end{table*}

Experimental results (Table~\ref{tab:experimental_results}) show that ADB \cite{Zhang:21} and DA-ADB \cite{DA-ADB} are not robust and exhibit poor performance in the compositionally diverse subsets: Banking\_CG, OOS\_CG, and StackOverflow\_CG. These subsets are derived from more extensive datasets, namely Banking, OOS, and StackOverflow. This indicates that these models struggle to achieve compositional generalization in more challenging contexts.

Interestingly, ADB is found to be more robust than DA-ADB, particularly in the OOS\_CG subset, where the model has to predict a larger number of intents ($151$ intents) in the test phase. This is about twice the number of intents in Banking\_CG and $7.5$ times that of StackOverflow\_CG. However, DA-ADB outperforms ADB in the StackOverflow\_CG subset, which is more balanced and has far fewer intents to predict.

In the Banking\_CG subset, it was observed that the overall F1 scores of ADB with ChatGPT data augmentation were consistently higher (by about $2$ to $4\%$) than those of ADB and DA-ADB. A similar trend was seen in the OOS\_CG subset, where the F1 scores of ADB with ChatGPT data augmentation were $2$ to $5\%$ better than ADB and DA-ADB. These results demonstrate that data augmentation can indeed help bridge the gap between the training and test sets, even when they exhibit compositional dissimilarity.

ADB with ChatGPT data augmentation outperforms DA-ADB with augmentation in both Banking\_CG and OOS\_CG. Interestingly, GPTAUG-WP10, a more sophisticated data augmentation method (which paraphrases wrongly predicted instances), underperforms when compared to simply incorporating all ChatGPT paraphrases into the training process (GPTAUG-F4 and GPTAUG-F10).

Finally, DA-ADB performs best in StackOverflow\_CG, considering that this subset is relatively more balanced and has fewer intents to predict.

\section{Conclusion}
\label{sec:conclusion}
In conclusion, this paper addresses the challenge of compositional generalization in open intent detection by leveraging the capabilities of ChatGPT, a state-of-the-art large language model. By constructing compositionally diverse datasets (i.e., \textbf{Banking\_CG}, \textbf{OOS\_CG}, and \textbf{StackOverflow\_CG}) and incorporating ChatGPT-generated paraphrases into the training process, we have demonstrated large improvements in model performance on unseen compositions. 

Future research should focus on developing more advanced data augmentation approaches that can generate more diverse compositions. One possible direction involves designing better-instructed prompts for ChatGPT to encourage more diverse paraphrases that can help improve compositional generalization even further. Additionally, exploring alternative strategies for incorporating augmented data and refining the iterative training process may lead to further performance improvements. 

\newpage


\newpage

\appendix

\section{Dataset Construction in Detail}
\label{sec:appendix_dataset_construction_in_detail}

Banking\_CG, OOS\_CG, and StackOverflow\_CG are subsets derived from Banking, OOS, and StackOverflow by pruning compositionally-similar pairs of utterance instances between their training and test/development sets. Rouge-L score is adopted to identify overlap common subsequences in a pair of utterance instances. The larger Rouge-L score usually indicates that more common compositions (n-grams) are shared among the pair of utterances. 

In Banking\_CG and OOS\_CG, we used a Rouge-L threshold of $0.3$ to detect a similar pair of utterance instances between their training and test/development sets, while in StackOverflow\_CG, a threshold of $0.2$ is adopted. (Refer to Appendix~\ref{sec:appendix_examples_of_rouge_score_and_corresponding_utterane_pairs} for more about Rouge score and the corresponding utterance pairs.)

Once the compositionally-similar pair of utterance instances are identified, a graph is created by connecting these instances with edges, then highest-degree nodes (instances) and their edges are pruned iteratively. Considering that training and test/development sets had significantly different numbers of instances, the node degree is multiplied by the weight (the number of remaining instances in the set) to readjust if the node to prune should be from the training or test/development set.

The pruning process iterates until a certain condition is met. In Banking\_CG and OOS\_CG, we stopped the process when the maximum node degree of the test/development sets reached a number of $5$, while in StackOverflow\_CG, the process wasn't stopped until all similar pairs were pruned, considering that StackOverflow\_CG is relatively more balanced and has fewer intents to predict. 

In Banking\_CG, $6231$, $183$, and $1184$ utterance instances were pruned from the corresponding training, development and test sets, while in OOS\_CG, $11317$, $1306$, and $2068$ were pruned, and in StackOverflow\_CG, $9209$, $578$, and $3095$ instances were removed from their training, development and test sets, respectively. Detailed statistics of Banking\_CG, OOS\_CG, and StackOverflow\_CG can be found in Tables~\ref{tab:banking_cg_train_dataset_statistics} to \ref{tab:stackoverflow_cg_test_dataset_statistics}.

\begin{table*}
\centering
\small
\caption{\textbf{Banking\_CG} training dataset statistics}
\label{tab:banking_cg_train_dataset_statistics}
\begin{tabular}{||l| r||l|r||}
\hline
\textbf{Intent}&\textbf{\#Instance} & \textbf{Intent}&\textbf{\#Instance}\\
\hline
Refund\_not\_showing\_up&54&get\_physical\_card&18\\
activate\_my\_card&49&getting\_spare\_card&29\\
age\_limit&31&getting\_virtual\_card&16\\
apple\_pay\_or\_google\_pay&28&lost\_or\_stolen\_card&22\\
atm\_support&25&lost\_or\_stolen\_phone&24\\
automatic\_top\_up&31&order\_physical\_card&29\\
balance\_not\_updated\_after\_bank\_transfer&65&passcode\_forgotten&26\\
balance\_not\_updated\_after\_cheque\_or\_cash\_deposit&69&pending\_card\_payment&51\\
beneficiary\_not\_allowed&37&pending\_cash\_withdrawal&41\\
cancel\_transfer&49&pending\_top\_up&44\\
card\_about\_to\_expire&29&pending\_transfer&45\\
card\_acceptance&13&pin\_blocked&33\\
card\_arrival&51&receiving\_money&22\\
card\_delivery\_estimate&32&request\_refund&49\\
card\_linking&34&reverted\_card\_payment?&45\\
card\_not\_working&33&supported\_cards\_and\_currencies&38\\
card\_payment\_fee\_charged&85&terminate\_account&34\\
card\_payment\_not\_recognised&64&top\_up\_by\_bank\_transfer\_charge&20\\
card\_payment\_wrong\_exchange\_rate&68&top\_up\_by\_card\_charge&20\\
card\_swallowed&10&top\_up\_by\_cash\_or\_cheque&21\\
cash\_withdrawal\_charge&60&top\_up\_failed&35\\
cash\_withdrawal\_not\_recognised&43&top\_up\_limits&21\\
change\_pin&26&top\_up\_reverted&39\\
compromised\_card&14&topping\_up\_by\_card&20\\
contactless\_not\_working&15&transaction\_charged\_twice&57\\
country\_support&34&transfer\_fee\_charged&55\\
declined\_card\_payment&40&transfer\_into\_account&21\\
declined\_cash\_withdrawal&54&transfer\_not\_received\_by\_recipient&63\\
declined\_transfer&44&transfer\_timing&26\\
direct\_debit\_payment\_not\_recognised&89&unable\_to\_verify\_identity&15\\
disposable\_card\_limits&25&verify\_my\_identity&27\\
edit\_personal\_details&24&verify\_source\_of\_funds&22\\
exchange\_charge&25&verify\_top\_up&24\\
exchange\_rate&21&virtual\_card\_not\_working&5\\
exchange\_via\_app&24&visa\_or\_mastercard&37\\
extra\_charge\_on\_statement&51&why\_verify\_identity&30\\
failed\_transfer&33&wrong\_amount\_of\_cash\_received&67\\
fiat\_currency\_support&37&wrong\_exchange\_rate\_for\_cash\_withdrawal&53\\
get\_disposable\_virtual\_card&12&&\\
\hline
\end{tabular}
\end{table*}

\begin{table*}
\centering
\small
\caption{\textbf{Banking\_CG} development dataset statistics}
\label{tab:banking_cg_development_dataset_statistics}
\begin{tabular}{||l| r||l|r||}
\hline
\textbf{Intent}&\textbf{\#Instance} & \textbf{Intent}&\textbf{\#Instance}\\
\hline
Refund\_not\_showing\_up&13&get\_physical\_card&8\\
activate\_my\_card&12&getting\_spare\_card&10\\
age\_limit&9&getting\_virtual\_card&6\\
apple\_pay\_or\_google\_pay&9&lost\_or\_stolen\_card&6\\
atm\_support&9&lost\_or\_stolen\_phone&10\\
automatic\_top\_up&11&order\_physical\_card&9\\
balance\_not\_updated\_after\_bank\_transfer&14&passcode\_forgotten&5\\
balance\_not\_updated\_after\_cheque\_or\_cash\_deposit&15&pending\_card\_payment&15\\
beneficiary\_not\_allowed&15&pending\_cash\_withdrawal&11\\
cancel\_transfer&12&pending\_top\_up&13\\
card\_about\_to\_expire&10&pending\_transfer&14\\
card\_acceptance&4&pin\_blocked&7\\
card\_arrival&11&receiving\_money&8\\
card\_delivery\_estimate&8&request\_refund&17\\
card\_linking&10&reverted\_card\_payment?&14\\
card\_not\_working&7&supported\_cards\_and\_currencies&10\\
card\_payment\_fee\_charged&18&terminate\_account&9\\
card\_payment\_not\_recognised&17&top\_up\_by\_bank\_transfer\_charge&9\\
card\_payment\_wrong\_exchange\_rate&12&top\_up\_by\_card\_charge&8\\
card\_swallowed&4&top\_up\_by\_cash\_or\_cheque&10\\
cash\_withdrawal\_charge&16&top\_up\_failed&13\\
cash\_withdrawal\_not\_recognised&16&top\_up\_limits&8\\
change\_pin&11&top\_up\_reverted&14\\
compromised\_card&9&topping\_up\_by\_card&10\\
contactless\_not\_working&3&transaction\_charged\_twice&16\\
country\_support&12&transfer\_fee\_charged&17\\
declined\_card\_payment&14&transfer\_into\_account&8\\
declined\_cash\_withdrawal&17&transfer\_not\_received\_by\_recipient&14\\
declined\_transfer&10&transfer\_timing&9\\
direct\_debit\_payment\_not\_recognised&10&unable\_to\_verify\_identity&9\\
disposable\_card\_limits&8&verify\_my\_identity&7\\
edit\_personal\_details&7&verify\_source\_of\_funds&7\\
exchange\_charge&10&verify\_top\_up&11\\
exchange\_rate&7&virtual\_card\_not\_working&3\\
exchange\_via\_app&9&visa\_or\_mastercard&11\\
extra\_charge\_on\_statement&12&why\_verify\_identity&9\\
failed\_transfer&13&wrong\_amount\_of\_cash\_received&17\\
fiat\_currency\_support&9&wrong\_exchange\_rate\_for\_cash\_withdrawal&15\\
get\_disposable\_virtual\_card&7&&\\
\hline
\end{tabular}
\end{table*}

\begin{table*}
\centering
\small
\caption{\textbf{Banking\_CG} test dataset statistics}
\label{tab:banking_cg_test_dataset_statistics}
\begin{tabular}{||l| r||l|r||}
\hline
\textbf{Intent}&\textbf{\#Instance} & \textbf{Intent}&\textbf{\#Instance}\\
\hline
Refund\_not\_showing\_up&28&get\_physical\_card&18\\
activate\_my\_card&31&getting\_spare\_card&29\\
age\_limit&23&getting\_virtual\_card&21\\
apple\_pay\_or\_google\_pay&22&lost\_or\_stolen\_card&16\\
atm\_support&20&lost\_or\_stolen\_phone&24\\
automatic\_top\_up&24&order\_physical\_card&15\\
balance\_not\_updated\_after\_bank\_transfer&25&passcode\_forgotten&16\\
balance\_not\_updated\_after\_cheque\_or\_cash\_deposit&30&pending\_card\_payment&26\\
beneficiary\_not\_allowed&37&pending\_cash\_withdrawal&29\\
cancel\_transfer&29&pending\_top\_up&25\\
card\_about\_to\_expire&31&pending\_transfer&34\\
card\_acceptance&20&pin\_blocked&21\\
card\_arrival&24&receiving\_money&26\\
card\_delivery\_estimate&23&request\_refund&36\\
card\_linking&28&reverted\_card\_payment?&35\\
card\_not\_working&20&supported\_cards\_and\_currencies&24\\
card\_payment\_fee\_charged&17&terminate\_account&11\\
card\_payment\_not\_recognised&20&top\_up\_by\_bank\_transfer\_charge&20\\
card\_payment\_wrong\_exchange\_rate&20&top\_up\_by\_card\_charge&19\\
card\_swallowed&17&top\_up\_by\_cash\_or\_cheque&31\\
cash\_withdrawal\_charge&34&top\_up\_failed&30\\
cash\_withdrawal\_not\_recognised&34&top\_up\_limits&19\\
change\_pin&18&top\_up\_reverted&27\\
compromised\_card&18&topping\_up\_by\_card&17\\
contactless\_not\_working&20&transaction\_charged\_twice&35\\
country\_support&18&transfer\_fee\_charged&33\\
declined\_card\_payment&32&transfer\_into\_account&27\\
declined\_cash\_withdrawal&35&transfer\_not\_received\_by\_recipient&27\\
declined\_transfer&26&transfer\_timing&26\\
direct\_debit\_payment\_not\_recognised&16&unable\_to\_verify\_identity&30\\
disposable\_card\_limits&21&verify\_my\_identity&21\\
edit\_personal\_details&27&verify\_source\_of\_funds&26\\
exchange\_charge&24&verify\_top\_up&29\\
exchange\_rate&22&virtual\_card\_not\_working&9\\
exchange\_via\_app&22&visa\_or\_mastercard&20\\
extra\_charge\_on\_statement&36&why\_verify\_identity&22\\
failed\_transfer&27&wrong\_amount\_of\_cash\_received&29\\
fiat\_currency\_support&23&wrong\_exchange\_rate\_for\_cash\_withdrawal&28\\
get\_disposable\_virtual\_card&23&&\\
\hline
\end{tabular}
\end{table*}

\begin{table*}
\centering
\small
\caption{\textbf{OOS\_CG} training dataset statistics}
\label{tab:oos_cg_train_dataset_statistics}
\begin{tabular}{||l| r||l|r||l|r||}
\hline
\textbf{Intent}&\textbf{\#Instance} & \textbf{Intent}&\textbf{\#Instance} & \textbf{Intent}&\textbf{\#Instance}\\
\hline
accept\_reservations&50&greeting&33&reset\_settings&6\\
account\_blocked&21&how\_busy&20&restaurant\_reservation&22\\
alarm&13&how\_old\_are\_you&47&restaurant\_reviews&47\\
application\_status&22&improve\_credit\_score&11&restaurant\_suggestion&23\\
apr&27&income&52&rewards\_balance&15\\
are\_you\_a\_bot&20&ingredient\_substitution&42&roll\_dice&11\\
balance&22&ingredients\_list&26&rollover\_401k&19\\
bill\_balance&18&insurance&29&routing&29\\
bill\_due&15&insurance\_change&26&schedule\_maintenance&19\\
book\_flight&19&interest\_rate&26&schedule\_meeting&21\\
book\_hotel&17&international\_fees&16&share\_location&24\\
calculator&53&international\_visa&33&shopping\_list&24\\
calendar&34&jump\_start&10&shopping\_list\_update&16\\
calendar\_update&17&last\_maintenance&14&smart\_home&19\\
calories&41&lost\_luggage&28&spelling&33\\
cancel&30&make\_call&20&spending\_history&17\\
cancel\_reservation&19&maybe&26&sync\_device&7\\
car\_rental&12&meal\_suggestion&29&taxes&24\\
card\_declined&16&meaning\_of\_life&17&tell\_joke&26\\
carry\_on&32&measurement\_conversion&23&text&27\\
change\_accent&23&meeting\_schedule&45&thank\_you&25\\
change\_ai\_name&15&min\_payment&23&time&29\\
change\_language&33&mpg&26&timer&14\\
change\_speed&27&new\_card&24&timezone&39\\
change\_user\_name&47&next\_holiday&17&tire\_change&26\\
change\_volume&11&next\_song&24&tire\_pressure&21\\
confirm\_reservation&25&no&25&todo\_list&12\\
cook\_time&24&nutrition\_info&28&todo\_list\_update&21\\
credit\_limit&17&oil\_change\_how&17&traffic&16\\
credit\_limit\_change&13&oil\_change\_when&15&transactions&26\\
credit\_score&6&order&49&transfer&17\\
current\_location&18&order\_checks&20&translate&24\\
damaged\_card&17&order\_status&21&travel\_alert&55\\
date&31&pay\_bill&23&travel\_notification&27\\
definition&62&payday&22&travel\_suggestion&29\\
direct\_deposit&14&pin\_change&10&uber&14\\
directions&40&play\_music&42&update\_playlist&27\\
distance&57&plug\_type&12&user\_name&23\\
do\_you\_have\_pets&19&pto\_balance&7&vaccines&23\\
exchange\_rate&33&pto\_request&35&w2&18\\
expiration\_date&13&pto\_request\_status&18&weather&21\\
find\_phone&11&pto\_used&14&what\_are\_your\_hobbies&26\\
flight\_status&29&recipe&37&what\_can\_i\_ask\_you&6\\
flip\_coin&13&redeem\_rewards&19&what\_is\_your\_name&26\\
food\_last&43&reminder&52&what\_song&28\\
freeze\_account&23&reminder\_update&26&where\_are\_you\_from&23\\
fun\_fact&18&repeat&14&whisper\_mode&23\\
gas&13&replacement\_card\_duration&16&who\_do\_you\_work\_for&32\\
gas\_type&12&report\_fraud&11&who\_made\_you&43\\
goodbye&43&report\_lost\_card&25&yes&47\\
\hline
\end{tabular}
\end{table*}

\begin{table*}
\centering
\small
\caption{\textbf{OOS\_CG} development dataset statistics}
\label{tab:oos_cg_development_dataset_statistics}
\begin{tabular}{||l| r||l|r||l|r||}
\hline
\textbf{Intent}&\textbf{\#Instance} & \textbf{Intent}&\textbf{\#Instance} & \textbf{Intent}&\textbf{\#Instance}\\
\hline
accept\_reservations&15&greeting&13&reset\_settings&13\\
account\_blocked&11&how\_busy&15&restaurant\_reservation&10\\
alarm&14&how\_old\_are\_you&11&restaurant\_reviews&9\\
application\_status&12&improve\_credit\_score&9&restaurant\_suggestion&19\\
apr&6&income&6&rewards\_balance&9\\
are\_you\_a\_bot&10&ingredient\_substitution&13&roll\_dice&13\\
balance&14&ingredients\_list&12&rollover\_401k&9\\
bill\_balance&11&insurance&11&routing&4\\
bill\_due&9&insurance\_change&9&schedule\_maintenance&16\\
book\_flight&18&interest\_rate&9&schedule\_meeting&8\\
book\_hotel&15&international\_fees&15&share\_location&14\\
calculator&10&international\_visa&5&shopping\_list&5\\
calendar&9&jump\_start&17&shopping\_list\_update&13\\
calendar\_update&16&last\_maintenance&11&smart\_home&18\\
calories&5&lost\_luggage&15&spelling&11\\
cancel&18&make\_call&14&spending\_history&16\\
cancel\_reservation&19&maybe&18&sync\_device&13\\
car\_rental&11&meal\_suggestion&14&taxes&11\\
card\_declined&11&meaning\_of\_life&15&tell\_joke&13\\
carry\_on&18&measurement\_conversion&15&text&9\\
change\_accent&15&meeting\_schedule&8&thank\_you&15\\
change\_ai\_name&13&min\_payment&8&time&3\\
change\_language&8&mpg&12&timer&14\\
change\_speed&14&new\_card&7&timezone&8\\
change\_user\_name&14&next\_holiday&11&tire\_change&8\\
change\_volume&15&next\_song&9&tire\_pressure&11\\
confirm\_reservation&12&no&15&todo\_list&12\\
cook\_time&9&nutrition\_info&11&todo\_list\_update&5\\
credit\_limit&4&oil\_change\_how&4&traffic&11\\
credit\_limit\_change&12&oil\_change\_when&8&transactions&14\\
credit\_score&5&order&13&transfer&9\\
current\_location&11&order\_checks&16&translate&15\\
damaged\_card&11&order\_status&16&travel\_alert&11\\
date&10&pay\_bill&8&travel\_notification&12\\
definition&12&payday&9&travel\_suggestion&14\\
direct\_deposit&6&pin\_change&13&uber&11\\
directions&10&play\_music&16&update\_playlist&6\\
distance&6&plug\_type&11&user\_name&11\\
do\_you\_have\_pets&14&pto\_balance&9&vaccines&8\\
exchange\_rate&14&pto\_request&9&w2&11\\
expiration\_date&9&pto\_request\_status&10&weather&16\\
find\_phone&5&pto\_used&12&what\_are\_your\_hobbies&6\\
flight\_status&7&recipe&11&what\_can\_i\_ask\_you&12\\
flip\_coin&16&redeem\_rewards&6&what\_is\_your\_name&8\\
food\_last&11&reminder&9&what\_song&10\\
freeze\_account&4&reminder\_update&12&where\_are\_you\_from&19\\
fun\_fact&18&repeat&13&whisper\_mode&12\\
gas&10&replacement\_card\_duration&10&who\_do\_you\_work\_for&9\\
gas\_type&11&report\_fraud&14&who\_made\_you&9\\
goodbye&16&report\_lost\_card&8&yes&12\\
\hline
\end{tabular}
\end{table*}

\begin{table*}
\centering
\small
\caption{\textbf{OOS\_CG} test dataset statistics}
\label{tab:oos_cg_test_dataset_statistics}
\begin{tabular}{||l| r||l|r||l|r||}
\hline
\textbf{Intent}&\textbf{\#Instance} & \textbf{Intent}&\textbf{\#Instance} & \textbf{Intent}&\textbf{\#Instance}\\
\hline
accept\_reservations&17&how\_busy&22&restaurant\_reservation&17\\
account\_blocked&15&how\_old\_are\_you&16&restaurant\_reviews&20\\
alarm&30&improve\_credit\_score&8&restaurant\_suggestion&21\\
application\_status&19&income&17&rewards\_balance&12\\
apr&11&ingredient\_substitution&19&roll\_dice&17\\
are\_you\_a\_bot&17&ingredients\_list&20&rollover\_401k&11\\
balance&15&insurance&11&routing&5\\
bill\_balance&17&insurance\_change&11&schedule\_maintenance&12\\
bill\_due&10&interest\_rate&17&schedule\_meeting&11\\
book\_flight&15&international\_fees&22&share\_location&22\\
book\_hotel&17&international\_visa&6&shopping\_list&5\\
calculator&14&jump\_start&11&shopping\_list\_update&17\\
calendar&12&last\_maintenance&18&smart\_home&22\\
calendar\_update&19&lost\_luggage&24&spelling&15\\
calories&15&make\_call&20&spending\_history&14\\
cancel&23&maybe&20&sync\_device&12\\
cancel\_reservation&20&meal\_suggestion&17&taxes&16\\
car\_rental&15&meaning\_of\_life&23&tell\_joke&15\\
card\_declined&9&measurement\_conversion&21&text&18\\
carry\_on&22&meeting\_schedule&16&thank\_you&25\\
change\_accent&14&min\_payment&24&time&10\\
change\_ai\_name&18&mpg&25&timer&29\\
change\_language&17&new\_card&5&timezone&12\\
change\_speed&15&next\_holiday&21&tire\_change&8\\
change\_user\_name&14&next\_song&8&tire\_pressure&12\\
change\_volume&16&no&23&todo\_list&18\\
confirm\_reservation&14&nutrition\_info&25&todo\_list\_update&10\\
cook\_time&8&oil\_change\_how&9&traffic&13\\
credit\_limit&10&oil\_change\_when&11&transactions&20\\
credit\_limit\_change&21&oos&1200&transfer&15\\
credit\_score&28&order&17&translate&23\\
current\_location&21&order\_checks&17&travel\_alert&13\\
damaged\_card&16&order\_status&20&travel\_notification&14\\
date&15&pay\_bill&6&travel\_suggestion&15\\
definition&17&payday&13&uber&20\\
direct\_deposit&19&pin\_change&13&update\_playlist&8\\
directions&19&play\_music&21&user\_name&7\\
distance&15&plug\_type&18&vaccines&12\\
do\_you\_have\_pets&25&pto\_balance&9&w2&23\\
exchange\_rate&21&pto\_request&15&weather&26\\
expiration\_date&17&pto\_request\_status&15&what\_are\_your\_hobbies&13\\
find\_phone&9&pto\_used&20&what\_can\_i\_ask\_you&18\\
flight\_status&19&recipe&19&what\_is\_your\_name&15\\
flip\_coin&24&redeem\_rewards&19&what\_song&13\\
food\_last&16&reminder&14&where\_are\_you\_from&21\\
freeze\_account&9&reminder\_update&17&whisper\_mode&22\\
fun\_fact&11&repeat&22&who\_do\_you\_work\_for&13\\
gas&16&replacement\_card\_duration&11&who\_made\_you&12\\
gas\_type&15&report\_fraud&15&yes&16\\
goodbye&22&report\_lost\_card&15&&\\
greeting&21&reset\_settings&14&&\\
\hline
\end{tabular}
\end{table*}

\begin{table*}
\centering
\small
\caption{\textbf{StackOverflow\_CG} training dataset statistics}
\label{tab:stackoverflow_cg_train_dataset_statistics}
\begin{tabular}{||l| r||l|r||l|r||l|r||l|r||}
\hline
\textbf{Intent}&\textbf{\#Instance} & \textbf{Intent}&\textbf{\#Instance} & \textbf{Intent}&\textbf{\#Instance} & \textbf{Intent}&\textbf{\#Instance} & \textbf{Intent}&\textbf{\#Instance}\\
\hline
ajax&158&drupal&132&linq&105&osx&139&spring&154\\
apache&143&excel&106&magento&112&qt&147&svn&116\\
bash&104&haskell&126&matlab&132&scala&135&visual-studio&200\\
cocoa&233&hibernate&130&oracle&124&sharepoint&140&wordpress&155\\
\hline
\end{tabular}
\end{table*}

\begin{table*}
\centering
\small
\caption{\textbf{StackOverflow\_CG} development dataset statistics}
\label{tab:stackoverflow_cg_development_dataset_statistics}
\begin{tabular}{||l| r||l|r||l|r||l|r||l|r||}
\hline
\textbf{Intent}&\textbf{\#Instance} & \textbf{Intent}&\textbf{\#Instance} & \textbf{Intent}&\textbf{\#Instance} & \textbf{Intent}&\textbf{\#Instance} & \textbf{Intent}&\textbf{\#Instance}\\
\hline
ajax&74&drupal&70&linq&78&osx&91&spring&67\\
apache&80&excel&70&magento&66&qt&83&svn&75\\
bash&80&haskell&71&matlab&73&scala&72&visual-studio&42\\
cocoa&66&hibernate&68&oracle&63&sharepoint&75&wordpress&58\\
\hline
\end{tabular}
\end{table*}

\begin{table*}
\centering
\small
\caption{\textbf{StackOverflow\_CG} test dataset statistics}
\label{tab:stackoverflow_cg_test_dataset_statistics}
\begin{tabular}{||l| r||l|r||l|r||l|r||l|r||}
\hline
\textbf{Intent}&\textbf{\#Instance} & \textbf{Intent}&\textbf{\#Instance} & \textbf{Intent}&\textbf{\#Instance} & \textbf{Intent}&\textbf{\#Instance} & \textbf{Intent}&\textbf{\#Instance}\\
\hline
ajax&160&drupal&128&linq&131&osx&193&spring&151\\
apache&170&excel&140&magento&132&qt&172&svn&149\\
bash&139&haskell&122&matlab&164&scala&124&visual-studio&106\\
cocoa&161&hibernate&128&oracle&127&sharepoint&171&wordpress&137\\
\hline
\end{tabular}
\end{table*}

\section{Examples of Rouge Scores and Corresponding Utterance Pairs}
\label{sec:appendix_examples_of_rouge_score_and_corresponding_utterane_pairs}

The compositional similarity of a pair of utterances can be told by Rouge-L score. In the second row of Table~\ref{tab:examples_of_rouge_scores_and_corresponding_utterance_pairs}, given that the Rouge-L score is greater than $0.3$, a long span (4-gram) ``be using my card'' is shared by both the training and test utterance instances. When the Rouge-L score is not greater than $0.3$, the first row and the third row of Table~\ref{tab:examples_of_rouge_scores_and_corresponding_utterance_pairs}, literally those pairs are compositionally dissimilar and only a short span (bigram) ``my card'' is found common between training and test instances.

\begin{table*}
\centering
\small
\caption{Examples of Rouge scores and corresponding utterance pairs}
\label{tab:examples_of_rouge_scores_and_corresponding_utterance_pairs}
\begin{tabular}{|l|p{0.7\linewidth}|}
\thickhline
Training instance&Someone might be using my card that is not me.\\
Test instance&I don't recognize some of the transactions on my card, I think someone must have gotten my card info and used it.\\
Rouge-L score &0.13\\
\hline
Training instance&Someone might be using my card that is not me.\\
Test instance&What should I do if I think that someone else may be using my card.\\
Rouge-L score &0.33\\
\hline
Training instance&Someone might be using my card that is not me.\\
Test instance&I think someone got my card details and used it because there are transactions i don't recognize. What do I do now?\\
Rouge-L score &0.13\\
\thickhline
\end{tabular}
\end{table*}

\section{Examples of ChatGPT's Paraphrases}
\label{sec:appendix_examples_of_chatgpts_paraphrases}

Table~\ref{tab:examples_of_chatgpts_paraphrases} demonstrates that ChatGPT's paraphrases introduce diverse compositions from the original utterances.  For example, in the first row of Table~\ref{tab:examples_of_chatgpts_paraphrases_oos_cg_dataset}, the bigram ``equivalent of'' is replaced with a trigram ``corresponding phrase for'' of the same meaning. In the first row of Table\ref{tab:examples_of_chatgpts_paraphrases_banking_cg_dataset}, the original sentence ``i have a pending top-up'' is put into its passive voice structure. The diversities brought by ChatGPT's paraphrases eventually bridge the gap between compositionally dissimilar training and test sets.

\begin{table*}
\caption{Examples of ChatGPT's Paraphrases}
\label{tab:examples_of_chatgpts_paraphrases}
\begin{subtable}[h]{1\textwidth}
\centering
\small
\caption{Banking\_CG Dataset}
\label{tab:examples_of_chatgpts_paraphrases_banking_cg_dataset}
\begin{tabular}{|l|p{0.7\linewidth}|}
\thickhline
Original utterance&i have a pending top-up\\
ChatGPT's paraphrase&A top-up transaction is still being processed.\\
\hline
Original utterance&What are the currency types that I can keep money in.\\
ChatGPT's paraphrase&Which currencies are allowed for me to keep my finances?\\
\hline
Original utterance&I think my card is broken or something. I tried using it and it wouldn't go through, please help.\\
ChatGPT's paraphrase&It appears that my card may be broken or faulty, as I tried to use it but it didn't work. Could you provide me with some assistance, please?\\
\thickhline
\end{tabular}
\end{subtable}

\begin{subtable}[h]{1\textwidth}
\centering
\small
\caption{OOS\_CG Dataset}
\label{tab:examples_of_chatgpts_paraphrases_oos_cg_dataset}
\begin{tabular}{|l|p{0.7\linewidth}|}
\thickhline
Original utterance&what is the equivalent of, 'life is good' in french\\
ChatGPT's paraphrase&What is the corresponding phrase for 'life is good' in French?\\
\hline
Original utterance&how many inches and centimeters would i need if i had 10 feet of something\\
ChatGPT's paraphrase&How many inches and centimeters would be needed for 10 feet of an object?\\
\hline
Original utterance&please put taco night on the house calendar date of january 6\\
ChatGPT's paraphrase&Kindly mark January 6th as the date for taco night on the house calendar.\\
\thickhline
\end{tabular}
\end{subtable}

\begin{subtable}[h]{1\textwidth}
\centering
\small
\caption{StackOverflow\_CG Dataset}
\begin{tabular}{|l|p{0.7\linewidth}|}
\thickhline
Original utterance&Why this keyboard intercepting kernel extension doesn't work?\\
ChatGPT's paraphrase&Can you provide an answer for the non-functionality of the keyboard interception kernel extension?\\
\hline
Original utterance&NSColor with calibrated values works differently than regular color??\\
ChatGPT's paraphrase&Calibrated NSColor showcases a diverse behavior from a typical color.\\
\hline
Original utterance&Tips for using CVS or Subversion as a backup framework for office documents\\
ChatGPT's paraphrase&Guidelines for utilizing CVS or Subversion as a backup solution for office documents.\\
\thickhline
\end{tabular}
\end{subtable}
\end{table*}

\section{Experimental Results in Detail}
\label{sec:appendix_experimental_results_in_detail}
For a fair comparison, all settings are evaluated using the seed numbers $0$ to $9$ for known intent sampling. All settings are built on the BERT-base backbone and are optimized using the ADAM gradient descent algorithm. Full experimental results are shown in Tables~\ref{tab:adb_experimental_results_in_detail} to \ref{tab:da-adb_gptaug-wp10_experimental_results_in_detail}.

\begin{table*}
\centering
\small
\caption{ADB experimental results in detail}
\label{tab:adb_experimental_results_in_detail}
\begin{tabular}{|l|r|p{0.035\linewidth}p{0.035\linewidth}p{0.035\linewidth}p{0.035\linewidth}|p{0.035\linewidth}p{0.035\linewidth}p{0.035\linewidth}p{0.035\linewidth}|p{0.035\linewidth}p{0.035\linewidth}p{0.035\linewidth}p{0.035\linewidth}|}
\thickhline
&&\multicolumn{4}{|c|}{\bf Banking\_CG}&\multicolumn{4}{|c|}{\bf OOS\_CG}&\multicolumn{4}{|c|}{\bf StackOverflow\_CG}\\
\cline{3-14}
&\bf Seed&\bf F1-IND&\bf F1-OOD&\bf F1-All&\bf Acc-All&\bf F1-IND&\bf F1-OOD&\bf F1-All&\bf Acc-All&\bf F1-IND&\bf F1-OOD&\bf F1-All&\bf Acc-All\\
\hline
\multirow{9}{*}{25\%}&0&54.32&81.06&55.66&71.10&45.65&91.86&46.83&85.63&66.48&88.37&70.13&82.55\\
&1&54.36&79.78&55.63&70.94&52.36&91.18&53.35&84.77&60.63&80.47&63.94&72.94\\
&2&53.99&79.04&55.24&71.10&49.16&91.46&50.24&84.64&59.12&80.55&62.69&72.87\\
&3&53.55&80.15&54.88&71.78&49.81&89.61&50.83&82.57&59.95&85.19&64.16&78.04\\
&4&59.76&81.31&60.84&73.31&54.07&91.97&55.04&85.68&58.86&78.74&62.17&70.67\\
&5&49.72&81.85&51.33&72.52&50.03&90.73&51.07&84.03&55.42&69.00&57.68&61.17\\
&6&48.92&78.28&50.39&68.88&55.00&89.41&55.88&82.74&59.18&81.67&62.93&73.22\\
&7&50.79&83.28&52.41&74.21&44.05&89.56&45.22&82.46&61.03&84.09&64.88&76.90\\
&8&52.18&83.82&53.76&75.11&40.09&89.34&41.35&81.75&53.36&81.30&58.02&72.60\\
&9&57.33&82.40&58.58&74.10&51.07&91.41&52.11&85.32&49.71&61.01&51.59&52.56\\
\hline
\multirow{9}{*}{50\%}&0&57.44&72.01&57.82&66.51&51.47&83.75&51.89&75.55&75.71&82.67&76.34&80.07\\
&1&62.04&68.92&62.22&65.72&52.52&83.81&52.93&75.30&70.56&73.05&70.78&70.60\\
&2&61.66&70.50&61.89&66.93&50.64&85.10&51.09&76.29&70.76&78.05&71.42&74.18\\
&3&58.88&69.09&59.14&64.98&54.26&84.22&54.65&76.35&70.45&74.29&70.80&71.70\\
&4&62.11&69.96&62.31&66.30&55.06&84.01&55.44&75.63&68.29&71.15&68.55&69.12\\
&5&58.56&72.02&58.91&66.30&52.78&84.17&53.20&75.69&74.22&77.70&74.54&76.01\\
&6&55.92&67.11&56.21&62.08&51.95&83.97&52.37&75.94&73.29&80.95&73.99&77.45\\
&7&57.74&66.71&57.97&61.87&49.92&83.73&50.37&74.83&70.24&75.19&70.69&72.70\\
&8&60.93&71.71&61.21&67.41&49.19&82.40&49.62&73.49&69.59&73.54&69.95&71.53\\
&9&63.98&68.29&64.09&65.66&55.37&84.78&55.75&77.53&71.41&74.80&71.72&72.46\\
\hline
\multirow{9}{*}{75\%}&0&65.47&56.31&65.31&64.50&53.11&75.08&53.30&66.82&77.73&68.20&77.13&74.97\\
&1&66.17&51.73&65.93&63.71&53.75&76.63&53.95&68.83&74.97&57.43&73.87&69.60\\
&2&64.73&52.93&64.53&63.08&52.09&76.01&52.30&67.21&75.13&65.37&74.52&71.94\\
&3&63.41&49.66&63.18&61.50&55.98&75.57&56.15&68.86&75.72&57.86&74.60&69.98\\
&4&64.99&52.81&64.79&63.24&56.38&76.64&56.56&69.49&76.82&61.84&75.88&72.87\\
&5&64.98&59.67&64.89&64.87&51.57&78.20&51.81&68.78&78.43&59.41&77.24&72.25\\
&6&62.19&53.26&62.04&60.86&53.56&76.88&53.77&68.89&74.44&58.53&73.44&68.74\\
&7&59.66&51.17&59.51&59.02&52.58&76.27&52.79&68.01&78.70&66.63&77.95&75.22\\
&8&65.18&54.89&65.00&63.87&52.86&74.40&53.05&66.33&75.25&57.61&74.15&69.81\\
&9&66.26&51.22&66.01&63.50&56.87&76.72&57.05&70.10&74.12&62.73&73.40&70.43\\
\thickhline
\end{tabular}
\end{table*}

\begin{table*}
\centering
\small
\caption{DA-ADB experimental results in detail}
\begin{tabular}{|l|r|p{0.035\linewidth}p{0.035\linewidth}p{0.035\linewidth}p{0.035\linewidth}|p{0.035\linewidth}p{0.035\linewidth}p{0.035\linewidth}p{0.035\linewidth}|p{0.035\linewidth}p{0.035\linewidth}p{0.035\linewidth}p{0.035\linewidth}|}
\thickhline
&&\multicolumn{4}{|c|}{\bf Banking\_CG}&\multicolumn{4}{|c|}{\bf OOS\_CG}&\multicolumn{4}{|c|}{\bf StackOverflow\_CG}\\
\cline{3-14}
&\bf Seed&\bf F1-IND&\bf F1-OOD&\bf F1-All&\bf Acc-All&\bf F1-IND&\bf F1-OOD&\bf F1-All&\bf Acc-All&\bf F1-IND&\bf F1-OOD&\bf F1-All&\bf Acc-All\\
\hline
\multirow{9}{*}{25\%}&0&51.66&88.27&53.49&80.01&33.93&92.54&35.43&86.62&70.98&90.81&74.29&85.96\\
&1&55.33&84.21&56.77&76.21&37.62&91.84&39.01&85.55&66.88&87.19&70.26&81.03\\
&2&50.86&84.62&52.55&77.16&35.91&91.44&37.33&84.91&60.42&82.93&64.17&74.84\\
&3&55.55&85.63&57.05&78.22&39.84&91.12&41.16&84.61&66.89&88.05&70.41&82.03\\
&4&58.68&86.10&60.05&79.06&45.06&92.47&46.27&86.81&60.39&84.17&64.35&76.76\\
&5&51.81&86.97&53.57&78.96&44.04&92.12&45.27&86.29&62.54&81.87&65.76&74.22\\
&6&46.76&83.06&48.57&74.47&44.73&91.67&45.93&85.71&63.63&87.01&67.52&80.07\\
&7&49.18&87.05&51.08&78.74&35.03&90.76&36.46&83.92&64.84&86.87&68.51&80.45\\
&8&54.86&87.38&56.49&80.33&30.67&91.08&32.22&84.33&52.25&83.31&57.43&74.63\\
&9&58.64&88.16&60.11&81.17&35.87&91.93&37.31&85.74&54.41&76.14&58.04&66.85\\
\hline
\multirow{9}{*}{50\%}&0&49.77&75.22&50.42&66.72&31.77&84.11&32.46&74.15&80.15&86.31&80.71&84.17\\
&1&52.47&71.43&52.96&65.19&34.71&83.17&35.34&73.24&77.32&81.37&77.69&78.93\\
&2&57.29&74.85&57.74&69.30&31.49&83.23&32.17&72.91&72.93&81.07&73.67&77.18\\
&3&53.55&72.26&54.03&66.14&34.22&83.41&34.86&73.82&74.07&77.89&74.42&75.52\\
&4&55.62&73.03&56.07&66.77&37.00&82.77&37.60&72.96&74.59&81.62&75.23&78.66\\
&5&52.39&76.15&53.00&68.57&37.80&83.32&38.40&73.68&75.76&80.38&76.18&78.18\\
&6&50.45&77.04&51.13&67.93&35.98&83.47&36.61&73.84&79.23&85.97&79.84&83.17\\
&7&55.90&74.81&56.38&68.04&32.06&83.22&32.73&72.80&76.72&83.76&77.36&81.31\\
&8&58.98&77.10&59.44&71.20&31.04&82.92&31.72&72.60&71.85&77.45&72.36&74.87\\
&9&59.34&72.63&59.68&67.83&30.50&83.46&31.19&73.68&77.05&81.71&77.47&79.38\\
\hline
\multirow{9}{*}{75\%}&0&54.14&56.52&54.18&58.97&28.11&72.94&28.51&60.71&81.00&71.38&80.40&78.00\\
&1&55.05&51.20&54.99&56.75&28.66&71.49&29.04&59.55&75.19&63.00&74.43&71.46\\
&2&54.38&49.80&54.30&55.85&29.51&72.10&29.89&60.27&77.36&67.09&76.72&74.04\\
&3&55.38&48.25&55.26&55.27&28.06&71.12&28.44&59.36&78.03&61.71&77.01&72.74\\
&4&55.04&48.45&54.93&55.49&30.00&71.20&30.37&59.00&81.17&69.09&80.41&77.80\\
&5&52.59&55.11&52.63&56.80&29.10&73.25&29.49&60.79&79.46&62.63&78.40&74.01\\
&6&54.04&56.19&54.07&58.02&30.41&71.54&30.77&59.80&76.96&62.10&76.04&71.57\\
&7&51.78&51.07&51.77&54.32&29.22&71.38&29.59&59.42&81.15&70.17&80.47&77.97\\
&8&55.68&55.04&55.67&58.23&30.14&70.19&30.50&58.73&77.66&62.87&76.74&72.94\\
&9&59.35&52.95&59.24&59.70&32.61&72.40&32.97&61.43&77.73&67.98&77.12&74.56\\
\thickhline
\end{tabular}
\end{table*}

\begin{table*}
\centering
\small
\caption{ADB+GPTAUG-F4 experimental results in detail}
\begin{tabular}{|l|r|p{0.035\linewidth}p{0.035\linewidth}p{0.035\linewidth}p{0.035\linewidth}|p{0.035\linewidth}p{0.035\linewidth}p{0.035\linewidth}p{0.035\linewidth}|p{0.035\linewidth}p{0.035\linewidth}p{0.035\linewidth}p{0.035\linewidth}|}
\thickhline
&&\multicolumn{4}{|c|}{\bf Banking\_CG}&\multicolumn{4}{|c|}{\bf OOS\_CG}&\multicolumn{4}{|c|}{\bf StackOverflow\_CG}\\
\cline{3-14}
&\bf Seed&\bf F1-IND&\bf F1-OOD&\bf F1-All&\bf Acc-All&\bf F1-IND&\bf F1-OOD&\bf F1-All&\bf Acc-All&\bf F1-IND&\bf F1-OOD&\bf F1-All&\bf Acc-All\\
\hline
\multirow{9}{*}{25\%}&0&58.04&85.27&59.40&77.48&52.90&92.93&53.92&87.78&66.07&87.09&69.58&80.90\\
&1&57.11&81.53&58.33&73.63&56.85&92.69&57.77&87.33&63.04&83.53&66.46&76.56\\
&2&59.83&83.40&61.01&76.53&54.62&92.73&55.60&87.22&62.06&83.26&65.59&76.11\\
&3&58.91&83.93&60.16&76.74&53.28&90.81&54.24&84.64&59.22&84.74&63.48&77.52\\
&4&61.81&82.72&62.86&74.79&56.91&92.21&57.82&86.48&58.39&78.63&61.76&70.50\\
&5&53.76&84.91&55.31&77.00&56.47&92.70&57.40&87.47&57.87&71.56&60.15&63.86\\
&6&49.84&78.68&51.28&69.57&59.21&91.51&60.04&86.10&63.44&86.24&67.24&79.14\\
&7&51.00&84.02&52.65&75.37&45.39&89.72&46.53&82.76&61.32&84.23&65.14&77.11\\
&8&57.43&84.96&58.81&77.37&44.15&91.12&45.35&84.64&59.58&86.30&64.03&79.14\\
&9&59.54&84.31&60.78&76.58&55.61&92.91&56.56&87.89&58.38&80.57&62.08&72.36\\
\hline
\multirow{9}{*}{50\%}&0&59.88&72.93&60.21&67.99&55.53&85.53&55.92&78.30&74.80&82.30&75.48&79.48\\
&1&64.52&71.24&64.69&68.72&56.29&84.85&56.67&77.56&73.32&79.95&73.93&76.63\\
&2&64.21&72.72&64.43&69.67&53.62&85.84&54.04&77.92&69.22&78.59&70.07&74.04\\
&3&60.36&71.55&60.65&67.77&55.85&85.28&56.24&78.19&70.77&76.59&71.30&73.43\\
&4&65.22&74.51&65.46&70.62&58.56&85.23&58.91&78.14&66.06&64.09&65.88&64.27\\
&5&60.60&75.58&60.98&70.15&55.21&85.54&55.60&77.97&73.83&77.12&74.13&75.32\\
&6&60.24&76.12&60.65&71.10&57.29&86.50&57.68&79.90&73.91&82.90&74.72&79.17\\
&7&61.76&72.72&62.04&68.04&54.12&85.37&54.53&77.75&71.47&76.16&71.90&73.94\\
&8&63.54&73.68&63.80&70.31&51.68&84.15&52.11&76.21&68.46&72.63&68.84&70.43\\
&9&65.21&70.97&65.35&67.99&55.47&85.59&55.87&78.77&73.98&79.95&74.53&77.01\\
\hline
\multirow{9}{*}{75\%}&0&68.56&57.90&68.38&67.09&57.04&78.70&57.23&71.92&77.32&67.80&76.72&74.32\\
&1&68.72&53.37&68.46&65.98&58.07&77.79&58.25&71.45&76.73&62.79&75.86&72.43\\
&2&65.97&55.15&65.79&64.66&54.59&77.05&54.79&69.60&73.05&63.29&72.44&69.57\\
&3&65.24&50.29&64.98&62.97&56.72&76.96&56.90&71.01&75.24&59.45&74.25&69.64\\
&4&64.58&51.19&64.36&62.39&58.43&77.49&58.60&71.56&76.35&61.50&75.42&72.01\\
&5&66.94&58.90&66.81&65.88&53.77&78.05&53.98&70.43&77.68&59.26&76.53&71.60\\
&6&64.75&56.67&64.61&63.98&53.65&76.91&53.86&69.71&73.96&57.14&72.91&67.78\\
&7&67.40&55.04&67.19&65.72&53.77&77.22&53.98&70.04&78.29&65.85&77.51&74.60\\
&8&66.09&54.49&65.89&64.08&55.79&76.08&55.97&69.77&75.00&56.65&73.86&68.95\\
&9&68.24&55.17&68.02&66.14&58.04&77.31&58.21&71.48&73.54&63.11&72.88&69.91\\
\thickhline
\end{tabular}
\end{table*}

\begin{table*}
\centering
\small
\caption{ADB+GPTAUG-F10 experimental results in detail}
\begin{tabular}{|l|r|p{0.035\linewidth}p{0.035\linewidth}p{0.035\linewidth}p{0.035\linewidth}|p{0.035\linewidth}p{0.035\linewidth}p{0.035\linewidth}p{0.035\linewidth}|p{0.035\linewidth}p{0.035\linewidth}p{0.035\linewidth}p{0.035\linewidth}|}
\thickhline
&&\multicolumn{4}{|c|}{\bf Banking\_CG}&\multicolumn{4}{|c|}{\bf OOS\_CG}&\multicolumn{4}{|c|}{\bf StackOverflow\_CG}\\
\cline{3-14}
&\bf Seed&\bf F1-IND&\bf F1-OOD&\bf F1-All&\bf Acc-All&\bf F1-IND&\bf F1-OOD&\bf F1-All&\bf Acc-All&\bf F1-IND&\bf F1-OOD&\bf F1-All&\bf Acc-All\\
\hline
\multirow{9}{*}{25\%}&0&58.71&86.00&60.07&78.59&53.03&92.85&54.05&87.67&66.19&86.48&69.58&80.17\\
&1&57.87&81.50&59.05&73.68&57.33&92.70&58.24&87.42&62.82&83.08&66.19&75.97\\
&2&60.23&82.65&61.35&75.74&55.26&92.90&56.22&87.50&58.46&78.12&61.74&70.19\\
&3&58.84&83.84&60.09&76.64&52.85&91.08&53.83&84.99&53.44&76.63&57.30&68.26\\
&4&63.74&85.02&64.80&78.01&58.23&92.79&59.11&87.56&58.66&80.21&62.25&72.22\\
&5&54.81&85.70&56.36&78.11&58.45&92.92&59.33&87.91&53.65&67.25&55.92&59.66\\
&6&50.23&79.56&51.70&70.73&59.61&91.42&60.43&85.90&58.26&81.54&62.14&73.18\\
&7&52.00&84.93&53.65&76.58&46.39&89.63&47.49&82.63&60.53&83.16&64.30&75.73\\
&8&57.18&85.03&58.57&77.43&45.55&91.60&46.73&85.46&58.85&85.62&63.31&78.18\\
&9&62.15&86.17&63.36&79.06&55.95&92.81&56.90&87.75&59.03&81.32&62.75&73.08\\
\hline
\multirow{9}{*}{50\%}&0&59.45&73.31&59.81&68.51&56.05&86.22&56.45&79.32&74.30&82.36&75.03&79.35\\
&1&64.61&72.94&64.82&70.09&57.44&84.34&57.79&76.98&71.94&78.66&72.55&75.25\\
&2&63.61&73.12&63.85&69.99&53.01&86.08&53.45&78.61&68.62&78.56&69.52&74.11\\
&3&60.49&71.53&60.77&67.77&55.27&85.53&55.67&78.58&69.86&76.45&70.45&73.05\\
&4&65.83&74.73&66.06&71.26&58.02&85.55&58.38&78.69&67.31&68.90&67.45&67.37\\
&5&59.64&74.90&60.03&69.67&56.20&86.01&56.59&78.83&73.22&78.02&73.66&75.80\\
&6&60.02&76.72&60.45&71.62&56.64&86.21&57.03&79.71&74.21&83.34&75.04&79.62\\
&7&61.61&72.19&61.88&67.62&54.29&85.35&54.70&77.75&71.85&80.04&72.59&76.83\\
&8&62.66&74.10&62.95&70.57&51.47&84.50&51.90&76.82&69.22&75.75&69.81&72.87\\
&9&64.91&68.75&65.01&66.51&55.60&85.85&56.00&79.13&69.23&73.57&69.62&70.95\\
\hline
\multirow{9}{*}{75\%}&0&68.27&57.05&68.08&66.51&56.60&78.20&56.80&71.50&77.85&68.49&77.27&74.91\\
&1&68.10&53.69&67.86&65.56&56.99&77.18&57.17&70.90&76.53&60.94&75.56&71.29\\
&2&65.71&56.59&65.56&64.98&53.54&76.59&53.74&69.71&72.94&62.67&72.30&69.36\\
&3&65.29&52.34&65.07&63.55&57.20&76.27&57.37&70.46&74.57&58.22&73.55&68.74\\
&4&64.32&50.80&64.09&62.24&58.62&77.76&58.79&72.00&76.26&62.65&75.41&72.08\\
&5&65.75&57.27&65.60&64.61&54.58&77.99&54.79&70.65&77.08&59.47&75.97&71.33\\
&6&64.39&54.56&64.22&63.13&52.86&76.82&53.07&69.63&73.89&59.24&72.98&68.40\\
&7&66.59&54.30&66.38&65.08&53.26&76.52&53.47&69.36&77.40&64.11&76.56&73.56\\
&8&64.78&54.95&64.62&63.66&54.11&75.50&54.30&69.27&75.57&56.72&74.40&69.12\\
&9&68.97&54.48&68.72&66.03&58.63&77.61&58.80&71.94&71.42&59.03&70.64&67.33\\
\thickhline
\end{tabular}
\end{table*}

\begin{table*}
\centering
\small
\caption{ADB+GPTAUG-WP10 experimental results in detail}
\begin{tabular}{|l|r|p{0.035\linewidth}p{0.035\linewidth}p{0.035\linewidth}p{0.035\linewidth}|p{0.035\linewidth}p{0.035\linewidth}p{0.035\linewidth}p{0.035\linewidth}|p{0.035\linewidth}p{0.035\linewidth}p{0.035\linewidth}p{0.035\linewidth}|}
\thickhline
&&\multicolumn{4}{|c|}{\bf Banking\_CG}&\multicolumn{4}{|c|}{\bf OOS\_CG}&\multicolumn{4}{|c|}{\bf StackOverflow\_CG}\\
\cline{3-14}
&\bf Seed&\bf F1-IND&\bf F1-OOD&\bf F1-All&\bf Acc-All&\bf F1-IND&\bf F1-OOD&\bf F1-All&\bf Acc-All&\bf F1-IND&\bf F1-OOD&\bf F1-All&\bf Acc-All\\
\hline
\multirow{9}{*}{25\%}&0&53.68&74.20&54.71&63.82&42.78&88.54&43.95&80.23&55.82&55.25&55.72&50.26\\
&1&51.63&71.00&52.59&61.60&46.34&86.13&47.36&77.18&52.86&61.73&54.34&55.11\\
&2&50.59&67.94&51.46&60.44&47.38&89.55&48.46&81.55&50.06&60.37&51.78&53.56\\
&3&54.12&74.17&55.12&65.98&49.64&87.59&50.62&79.71&49.37&70.08&52.83&61.55\\
&4&58.30&77.65&59.27&69.57&54.61&91.64&55.56&85.27&56.11&74.04&59.10&65.92\\
&5&45.41&72.00&46.74&61.81&50.45&90.10&51.47&83.20&53.80&68.79&56.30&61.27\\
&6&47.58&67.77&48.59&58.54&57.17&89.23&58.00&82.65&50.65&68.32&53.60&59.35\\
&7&42.20&54.62&42.82&47.20&42.85&85.61&43.95&76.57&50.25&62.95&52.37&55.59\\
&8&45.03&72.01&46.38&61.50&41.98&88.94&43.19&81.22&45.41&54.21&46.88&47.13\\
&9&51.87&73.33&52.94&64.24&47.13&88.34&48.19&80.70&51.82&59.69&53.13&51.91\\
\hline
\multirow{9}{*}{50\%}&0&58.94&62.37&59.03&60.13&52.23&82.91&52.64&74.45&71.75&74.91&72.04&73.29\\
&1&62.10&58.44&62.00&60.07&52.31&82.77&52.71&74.28&70.46&71.96&70.59&70.02\\
&2&61.63&65.70&61.73&63.61&51.46&84.34&51.89&75.39&64.61&65.84&64.73&64.20\\
&3&59.66&64.29&59.78&62.18&55.36&83.71&55.73&75.74&69.14&70.79&69.29&69.16\\
&4&62.01&59.84&61.95&60.50&56.44&83.40&56.79&74.72&56.88&29.67&54.41&46.92\\
&5&56.61&63.18&56.78&60.28&54.38&84.28&54.77&75.94&69.88&65.79&69.51&67.13\\
&6&56.27&58.66&56.33&57.49&53.58&83.42&53.97&75.33&62.22&55.86&61.64&58.42\\
&7&58.79&57.53&58.76&57.70&49.73&82.41&50.16&73.32&68.51&69.89&68.64&68.85\\
&8&60.05&61.67&60.09&60.44&49.75&80.34&50.15&71.23&65.71&64.05&65.56&64.54\\
&9&62.64&59.46&62.56&60.34&57.26&83.06&57.60&75.66&71.25&72.51&71.36&71.15\\
\hline
\multirow{9}{*}{75\%}&0&67.13&55.05&66.93&65.14&53.69&74.77&53.88&66.63&76.61&63.64&75.80&72.98\\
&1&66.55&45.06&66.19&62.45&55.60&75.61&55.78&67.84&74.09&50.07&72.59&67.37\\
&2&63.93&47.66&63.66&61.55&53.02&75.63&53.22&67.10&73.16&55.16&72.04&68.19\\
&3&65.11&46.54&64.80&62.13&57.03&75.30&57.19&68.69&72.96&39.88&70.89&64.89\\
&4&66.03&46.44&65.70&62.18&56.62&76.20&56.79&68.92&76.12&54.41&74.77&70.71\\
&5&66.35&55.35&66.16&64.35&52.77&77.94&52.99&68.45&74.51&43.38&72.56&67.02\\
&6&64.69&50.50&64.45&61.87&55.38&76.83&55.57&68.81&73.15&49.34&71.67&66.16\\
&7&61.94&46.28&61.68&60.07&52.85&75.50&53.06&67.37&75.28&54.81&74.01&70.22\\
&8&65.27&47.83&64.98&62.13&53.96&73.84&54.14&66.22&73.53&41.78&71.54&65.34\\
&9&65.17&39.13&64.73&60.34&58.19&76.18&58.35&69.66&69.21&46.78&67.81&63.79\\
\thickhline
\end{tabular}
\end{table*}

\begin{table*}
\centering
\small
\caption{DA-ADB+GPTAUG-F4 experimental results in detail}
\begin{tabular}{|l|r|p{0.035\linewidth}p{0.035\linewidth}p{0.035\linewidth}p{0.035\linewidth}|p{0.035\linewidth}p{0.035\linewidth}p{0.035\linewidth}p{0.035\linewidth}|p{0.035\linewidth}p{0.035\linewidth}p{0.035\linewidth}p{0.035\linewidth}|}
\thickhline
&&\multicolumn{4}{|c|}{\bf Banking\_CG}&\multicolumn{4}{|c|}{\bf OOS\_CG}&\multicolumn{4}{|c|}{\bf StackOverflow\_CG}\\
\cline{3-14}
&\bf Seed&\bf F1-IND&\bf F1-OOD&\bf F1-All&\bf Acc-All&\bf F1-IND&\bf F1-OOD&\bf F1-All&\bf Acc-All&\bf F1-IND&\bf F1-OOD&\bf F1-All&\bf Acc-All\\
\hline
\multirow{9}{*}{25\%}&0&55.09&86.53&56.66&78.69&38.42&92.36&39.80&86.43&71.49&90.20&74.61&85.16\\
&1&57.86&82.82&59.11&75.32&42.45&91.97&43.72&85.82&45.42&44.39&45.25&37.59\\
&2&50.66&83.60&52.31&76.11&41.85&91.92&43.14&85.85&63.07&84.41&66.63&77.04\\
&3&56.91&83.70&58.25&76.16&46.46&91.05&47.60&84.75&63.46&87.15&67.41&80.59\\
&4&62.07&84.82&63.21&77.74&50.34&92.73&51.42&87.47&55.97&80.22&60.01&71.64\\
&5&51.81&85.38&53.49&77.37&48.41&92.23&49.53&86.65&56.85&54.46&56.45&46.47\\
&6&47.89&83.49&49.67&75.16&53.37&92.06&54.36&86.65&68.30&88.74&71.71&82.65\\
&7&47.67&85.26&49.55&76.58&38.10&90.42&39.44&83.54&62.50&86.83&66.55&80.07\\
&8&56.29&85.71&57.76&78.22&36.07&91.53&37.50&85.19&54.72&84.43&59.68&76.28\\
&9&59.50&86.93&60.88&79.75&44.37&92.27&45.60&86.51&57.94&79.49&61.53&70.43\\
\hline
\multirow{9}{*}{50\%}&0&50.54&73.12&51.12&66.51&35.47&84.41&36.11&75.22&78.31&84.84&78.90&82.48\\
&1&59.47&75.31&59.88&70.83&40.56&83.88&41.13&74.92&54.14&13.38&50.43&37.56\\
&2&57.98&75.05&58.42&70.04&38.37&84.00&38.97&74.61&70.96&78.61&71.66&74.53\\
&3&55.67&72.29&56.10&67.25&38.04&83.84&38.65&74.81&70.70&75.70&71.15&72.87\\
&4&58.89&73.87&59.27&69.15&40.96&83.27&41.51&74.37&71.34&80.37&72.16&76.63\\
&5&54.31&74.02&54.81&68.09&39.35&83.48&39.93&74.26&76.41&80.97&76.82&79.00\\
&6&56.25&77.70&56.80&71.99&43.06&84.56&43.60&76.24&75.68&82.58&76.30&79.38\\
&7&55.99&74.21&56.46&68.04&37.55&84.02&38.16&74.56&75.92&83.00&76.56&80.34\\
&8&59.88&76.63&60.31&72.10&38.05&83.97&38.65&74.86&73.36&79.61&73.93&76.97\\
&9&61.65&74.48&61.98&70.09&37.13&84.52&37.75&75.74&75.95&81.64&76.46&79.04\\
\hline
\multirow{9}{*}{75\%}&0&55.25&54.49&55.24&58.97&33.04&73.38&33.40&62.42&80.62&71.23&80.03&77.73\\
&1&57.69&50.92&57.57&58.18&33.93&72.34&34.27&61.70&78.27&65.96&77.50&74.32\\
&2&55.24&50.60&55.16&56.96&35.07&72.62&35.41&62.33&76.72&65.31&76.00&72.60\\
&3&53.68&46.91&53.57&54.17&32.50&72.04&32.85&61.59&77.15&60.04&76.08&71.33\\
&4&53.99&47.43&53.88&54.96&34.23&72.39&34.57&62.25&78.95&65.26&78.09&74.94\\
&5&54.30&54.09&54.30&57.81&31.94&73.10&32.30&61.59&78.56&61.99&77.52&73.05\\
&6&53.14&52.15&53.12&56.70&34.54&72.64&34.87&62.25&75.95&60.29&74.97&70.15\\
&7&57.41&52.71&57.33&58.91&32.39&71.39&32.73&60.57&73.60&50.75&72.17&66.57\\
&8&56.93&54.07&56.88&59.39&31.40&70.80&31.75&60.05&75.43&49.51&73.81&67.23\\
&9&61.08&52.53&60.94&60.60&35.98&73.35&36.31&63.33&73.46&57.54&72.47&67.13\\
\thickhline
\end{tabular}
\end{table*}

\begin{table*}
\centering
\small
\caption{DA-ADB+GPTAUG-F10 experimental results in detail}
\begin{tabular}{|l|r|p{0.035\linewidth}p{0.035\linewidth}p{0.035\linewidth}p{0.035\linewidth}|p{0.035\linewidth}p{0.035\linewidth}p{0.035\linewidth}p{0.035\linewidth}|p{0.035\linewidth}p{0.035\linewidth}p{0.035\linewidth}p{0.035\linewidth}|}
\thickhline
&&\multicolumn{4}{|c|}{\bf Banking\_CG}&\multicolumn{4}{|c|}{\bf OOS\_CG}&\multicolumn{4}{|c|}{\bf StackOverflow\_CG}\\
\cline{3-14}
&\bf Seed&\bf F1-IND&\bf F1-OOD&\bf F1-All&\bf Acc-All&\bf F1-IND&\bf F1-OOD&\bf F1-All&\bf Acc-All&\bf F1-IND&\bf F1-OOD&\bf F1-All&\bf Acc-All\\
\hline
\multirow{9}{*}{25\%}&0&52.79&85.62&54.43&77.58&38.97&92.28&40.34&86.29&65.05&79.99&67.54&71.88\\
&1&56.34&81.59&57.61&73.73&45.12&92.08&46.32&86.07&58.02&81.47&61.93&72.91\\
&2&50.23&82.99&51.87&75.32&41.15&91.85&42.45&85.66&54.62&76.51&58.26&66.33\\
&3&57.28&83.52&58.59&75.95&44.85&90.85&46.03&84.39&70.38&90.84&73.79&85.78\\
&4&56.53&82.43&57.83&73.95&50.29&92.19&51.36&86.59&59.89&84.56&64.00&76.97\\
&5&52.06&85.16&53.72&77.06&46.97&92.03&48.13&86.32&58.56&75.63&61.41&66.54\\
&6&47.94&83.57&49.72&75.26&51.96&91.89&52.99&86.32&71.34&90.61&74.55&85.40\\
&7&49.23&84.65&51.00&76.00&37.52&90.29&38.87&83.31&63.71&88.52&67.84&82.34\\
&8&55.30&84.64&56.76&76.90&38.27&91.38&39.63&85.02&56.27&84.82&61.03&76.63\\
&9&57.45&85.85&58.87&78.22&46.91&92.54&48.08&87.00&40.31&0.17&33.62&16.18\\
\hline
\multirow{9}{*}{50\%}&0&50.07&71.87&50.63&65.45&36.62&84.47&37.25&75.36&77.45&85.47&78.18&82.86\\
&1&58.32&75.07&58.75&70.46&39.97&83.92&40.55&74.86&55.61&27.05&53.01&42.10\\
&2&59.39&74.99&59.79&70.46&38.79&84.03&39.39&74.64&70.36&80.53&71.29&76.21\\
&3&53.08&71.80&53.56&66.30&38.38&83.62&38.98&74.56&67.74&76.56&68.54&72.74\\
&4&56.98&73.86&57.41&68.83&40.59&83.27&41.15&74.34&70.46&77.36&71.09&73.80\\
&5&53.81&74.10&54.33&67.99&40.70&83.56&41.26&74.42&75.73&81.43&76.25&79.00\\
&6&56.50&77.59&57.04&71.99&42.66&84.53&43.21&76.16&73.76&83.17&74.61&79.52\\
&7&55.97&74.49&56.44&68.30&39.05&83.96&39.64&74.70&71.88&81.52&72.76&78.18\\
&8&59.42&76.05&59.85&71.15&36.42&83.68&37.04&74.34&71.55&78.52&72.18&75.59\\
&9&61.70&74.40&62.03&69.99&37.01&84.37&37.63&75.58&68.63&76.20&69.32&72.67\\
\hline
\multirow{9}{*}{75\%}&0&53.43&53.49&53.43&57.54&34.01&73.64&34.36&63.33&77.10&67.88&76.52&74.25\\
&1&56.61&49.72&56.49&57.01&34.39&72.33&34.73&61.98&76.49&64.84&75.76&72.91\\
&2&54.23&50.28&54.16&56.17&36.57&73.04&36.89&63.24&74.02&63.03&73.34&69.95\\
&3&53.09&45.94&52.97&53.43&31.37&71.57&31.73&60.66&72.03&52.61&70.81&65.47\\
&4&53.99&46.46&53.86&54.22&33.32&72.17&33.66&61.95&73.55&54.27&72.35&68.50\\
&5&54.03&53.96&54.02&57.54&32.15&73.15&32.52&61.81&76.89&60.00&75.83&71.19\\
&6&52.23&52.07&52.22&56.17&34.82&72.76&35.15&62.39&73.87&57.40&72.84&67.54\\
&7&54.64&50.03&54.56&56.01&33.02&71.70&33.36&60.96&77.18&67.03&76.54&74.35\\
&8&56.11&52.99&56.06&58.44&32.94&71.29&33.28&60.93&71.23&51.46&70.00&64.75\\
&9&60.22&51.66&60.07&59.92&35.47&73.15&35.80&63.19&64.18&9.64&60.77&52.05\\
\thickhline
\end{tabular}
\end{table*}

\begin{table*}
\centering
\small
\caption{DA-ADB+GPTAUG-WP10 experimental results in detail}
\label{tab:da-adb_gptaug-wp10_experimental_results_in_detail}
\begin{tabular}{|l|r|p{0.035\linewidth}p{0.035\linewidth}p{0.035\linewidth}p{0.035\linewidth}|p{0.035\linewidth}p{0.035\linewidth}p{0.035\linewidth}p{0.035\linewidth}|p{0.035\linewidth}p{0.035\linewidth}p{0.035\linewidth}p{0.035\linewidth}|}
\thickhline
&&\multicolumn{4}{|c|}{\bf Banking\_CG}&\multicolumn{4}{|c|}{\bf OOS\_CG}&\multicolumn{4}{|c|}{\bf StackOverflow\_CG}\\
\cline{3-14}
&\bf Seed&\bf F1-IND&\bf F1-OOD&\bf F1-All&\bf Acc-All&\bf F1-IND&\bf F1-OOD&\bf F1-All&\bf Acc-All&\bf F1-IND&\bf F1-OOD&\bf F1-All&\bf Acc-All\\
\hline
\multirow{9}{*}{25\%}&0&53.70&86.98&55.37&78.06&38.21&92.80&39.61&87.00&63.39&81.83&66.46&74.73\\
&1&57.08&81.07&58.28&72.47&44.31&91.58&45.53&85.24&47.28&24.40&43.47&30.12\\
&2&55.92&81.19&57.18&73.58&39.26&91.75&40.61&85.41&47.71&29.83&44.73&31.57\\
&3&55.73&82.63&57.08&74.47&45.49&90.52&46.64&83.92&59.70&82.07&63.43&74.60\\
&4&62.46&83.43&63.51&75.79&50.07&92.43&51.15&86.81&58.29&77.75&61.53&69.74\\
&5&50.26&83.71&51.93&74.42&44.47&91.75&45.69&85.74&52.03&70.83&55.16&62.79\\
&6&49.43&77.93&50.85&68.78&49.18&91.79&50.28&86.10&60.47&80.97&63.88&73.22\\
&7&47.37&84.04&49.20&74.37&39.63&91.01&40.95&84.44&52.04&68.82&54.83&60.34\\
&8&55.20&85.40&56.71&77.74&37.16&90.51&38.53&83.65&47.95&60.58&50.06&52.32\\
&9&60.03&82.55&61.15&74.16&44.05&91.30&45.26&85.02&57.25&71.67&59.65&62.75\\
\hline
\multirow{9}{*}{50\%}&0&53.41&73.46&53.92&66.09&38.59&84.64&39.20&75.39&77.29&82.41&77.76&80.24\\
&1&62.07&73.60&62.37&69.30&41.71&83.77&42.27&74.45&64.02&27.95&60.75&43.58\\
&2&56.87&64.58&57.06&60.60&39.02&84.46&39.62&75.06&57.58&37.95&55.80&47.02\\
&3&59.00&70.99&59.31&66.09&41.87&83.77&42.42&74.89&68.68&67.71&68.59&67.13\\
&4&60.97&70.86&61.22&66.56&42.25&83.81&42.80&74.64&70.05&73.94&70.40&71.39\\
&5&57.57&71.68&57.93&65.24&41.63&83.77&42.19&74.72&64.25&40.44&62.09&50.57\\
&6&52.47&70.47&52.93&63.13&42.35&84.68&42.91&76.21&79.65&85.86&80.21&83.20\\
&7&58.65&70.89&58.96&65.61&38.17&83.97&38.78&74.26&75.57&82.13&76.17&79.59\\
&8&60.21&68.56&60.43&64.19&39.95&83.01&40.51&73.54&71.58&74.68&71.86&72.98\\
&9&65.06&68.23&65.14&66.19&37.06&83.97&37.68&74.97&70.44&72.75&70.65&70.81\\
\hline
\multirow{9}{*}{75\%}&0&61.38&56.01&61.29&62.34&38.23&74.93&38.56&64.34&81.59&71.79&80.98&78.62\\
&1&63.18&51.87&62.99&62.13&39.36&72.85&39.66&63.02&78.88&62.05&77.83&73.56\\
&2&61.95&52.63&61.80&61.18&40.57&74.43&40.87&64.26&77.74&65.95&77.01&73.91\\
&3&63.84&48.10&63.57&60.92&40.33&73.76&40.63&64.40&77.20&54.47&75.78&70.36\\
&4&62.31&49.46&62.10&60.86&41.52&74.05&41.80&64.45&80.47&65.99&79.56&76.49\\
&5&59.10&55.17&59.03&60.81&38.71&75.38&39.04&64.37&78.20&59.34&77.02&72.46\\
&6&60.86&56.98&60.80&61.81&40.03&74.08&40.33&64.04&77.30&60.53&76.25&71.81\\
&7&63.63&55.36&63.49&63.50&40.06&73.70&40.36&63.66&81.28&68.34&80.47&77.52\\
&8&62.06&55.62&61.95&62.24&39.52&71.95&39.81&62.31&70.73&37.47&68.65&60.83\\
&9&64.74&52.16&64.52&62.87&43.73&75.61&44.01&66.24&74.30&60.24&73.42&69.60\\
\thickhline
\end{tabular}
\end{table*}

\end{document}